# Phonetic based SoundEx & ShapeEx algorithm for Sindhi Spell Checker System

**[1]Zeeshan Bhatti, [2]Ahmad Waqas, [1]Imdad Ali Ismaili, [1]Dil Nawaz Hakro, [1]Waseem Javaid Soomro**

*[1]Institute of Information and Communication Technology, University of Sindh, Jamshoro, Pakistan.*
*[2]Department of Computer Science, Sukkur Institute of Business Administration, Sukkur, Pakistan.*

**ABSTRACT**

This paper presents a novel combinational phonetic algorithm for Sindhi Language, to be used in developing Sindhi Spell Checker which has yet not been developed prior to this work. The compound textual forms and glyphs of Sindhi language presents a substantial challenge for developing Sindhi spell checker system and generating similar suggestion list for misspelled words. In order to implement such a system, phonetic based Sindhi language rules and patterns must be considered into account for increasing the accuracy and efficiency. The proposed system is developed with a blend between Phonetic based SoundEx algorithm and ShapeEx algorithm for pattern or glyph matching, generating accurate and efficient suggestion list for incorrect or misspelled Sindhi words. A table of phonetically similar sounding Sindhi characters for SoundEx algorithm is also generated along with another table containing similar glyph or shape based character groups for ShapeEx algorithm. Both these are first ever attempt of any such type of categorization and representation for Sindhi Language.

## INTRODUCTION

Spell checkers are application programs that flag words in a document which may not be spelled correctly. The primary and most complex task of a spell checker system is to generate an accurate suggestion list for the misspelled word. While a significant amount of work has already been done for string matching algorithms in English and other Asian languages [1, 2, 3, 4, 5], on the contrary no work has been done for Sindhi Language. A study on trends in Sindhi text errors and patterns shows that the most common misspellings factors are the phonetically similar words with (i) similar sound/ pronunciation and (ii) similar shape/ glyphs of Sindhi Characters [6]. This gives the motivation to initially develop a phonetic based algorithm for Sindhi Text before a more practical Sindhi Spell checker can be developed. Therefore a compendium algorithm is developed; involving Edit Distance, Phonetic and Glyph/pattern matching algorithm to provide suggestions of similar words, for the misspelled word. This combinational hybrid approach successfully generates accurate suggestions for the misspelled Sindhi word.

SoundEx is generally considered as a phonetic algorithm, used primarily in natural Language Processing (NLP) for indexing names by sound. Simply stating, the SoundEx algorithm is used to group similar sounding letters together and assign each group a numerical number. The main goal of this technique is to use homophones for encoding text with numerical indexing; as a result the numerical representation can be easily matched with other similar sounding characters having same numerical code. This results in retrieving a list of words that are pronounced similarly with very little variation in their homophones. On the other hand the ShapeEx algorithm is used to group letters having same/similar shape or glyph and assign each group with a numerical number. Here the similar letters are those that are exactly same in shape and size, and are identified through certain diacritic marks such as ا{Alif} and آ{Alif-Mad-Aa}. These both letters are same except an additional symbol '~' is used above the آ. This distinguishes both in pronunciation, but typists usually confuse and make a common mistake of missing such type of characters as discussed in [6,7].

*Related Work:*

Phonetic algorithms for matching strings have been discussed in context to English language by various researchers from SoundEx [8,9], Metaphone [10,11] and Phonix [12]. Whereas some work in Sindhi Phonology and Phonetics has been done Jatoi A.N. as early as 1968 [13]. Sindhi morphological sound structure, vowel and consonant sounds, and its syntax have been discussed by [13,14,15,16]. However no work in developing a SoundEx and ShapeEx based algorithms for Sindhi was found in literature, thus making our effort first ever attempt in this area.

On the contrary a huge amount of work has been done in Arabic and other Asian languages in this context. Bird S., [17] presents a theory on Arabic Phonology based on Arabic verb morphology, syllable structure, and phonological constraints. Other work in Arabic Language Phonological and phonetic algorithms and structural analysis involves [18,19]. Recently Ousidhoumet et al. discusses the refinement of Arabic soundEx using two functions called "Algerian Dialect and Speech Therapy Refinement" [20]. Whereas Naushad-Uz-Zaman devolved

**Corresponding Author:** Zeeshan Bhatti, Institute of Information and Communication Technology, University of Sindh, Jamshoro, Pakistan.
Phone:+92 333 26066 30; E-mail: zeeshan.bhatti@usindh.edu.pk

a Spell checker using the Double Metaphone Encoding for Bangla language [21]. However Bal-Kirsna in his paper on "Nepali Spell Checker" has adopted an affix rules with pattern matching algorithm [22]. In contrast to this, T. Dhanabalan used Lexicons with morphological and syntactic information for the development of Tamil Spell Checker [23]. Tahira Naseem used various methodologies in the development of Urdu Spell checker such as Edit Distance algorithm, N-Grams, Sound-ex and its variants, using a Novel approach for ranking of spelling error corrections [24], and Probabilistic error correction techniques [25]. Rupinderdeep Kaur in his paper investigates various types of errors in Punjabi language with Gurmukhi Script and discusses various solutions and methodologies for creation of dictionary of Punjabi words, error detection and error correction & replacement [26].

The most relevant work on Sindhi phonology is done by Maharet at., where Sindhi Language phonology is discussed with "conversion of Sindhi alphabets letters into their appropriate sounds" [27].We have used the Shape groups discussed by Mahar et al. for similarly pronounced Sindhi phonemes and grouped them together to be used in our phonetic algorithms. We then created our own SoundEx based list of Sindhi character along with similar shape list for the ShapeEx algorithms. The available list of 360 Sindhi Phonemes as shown in figure 1; obtained from [27] and [28], is used to generate a basic list of similarly pronounced character list for phonetic based Sindhi SoundEx algorithm.

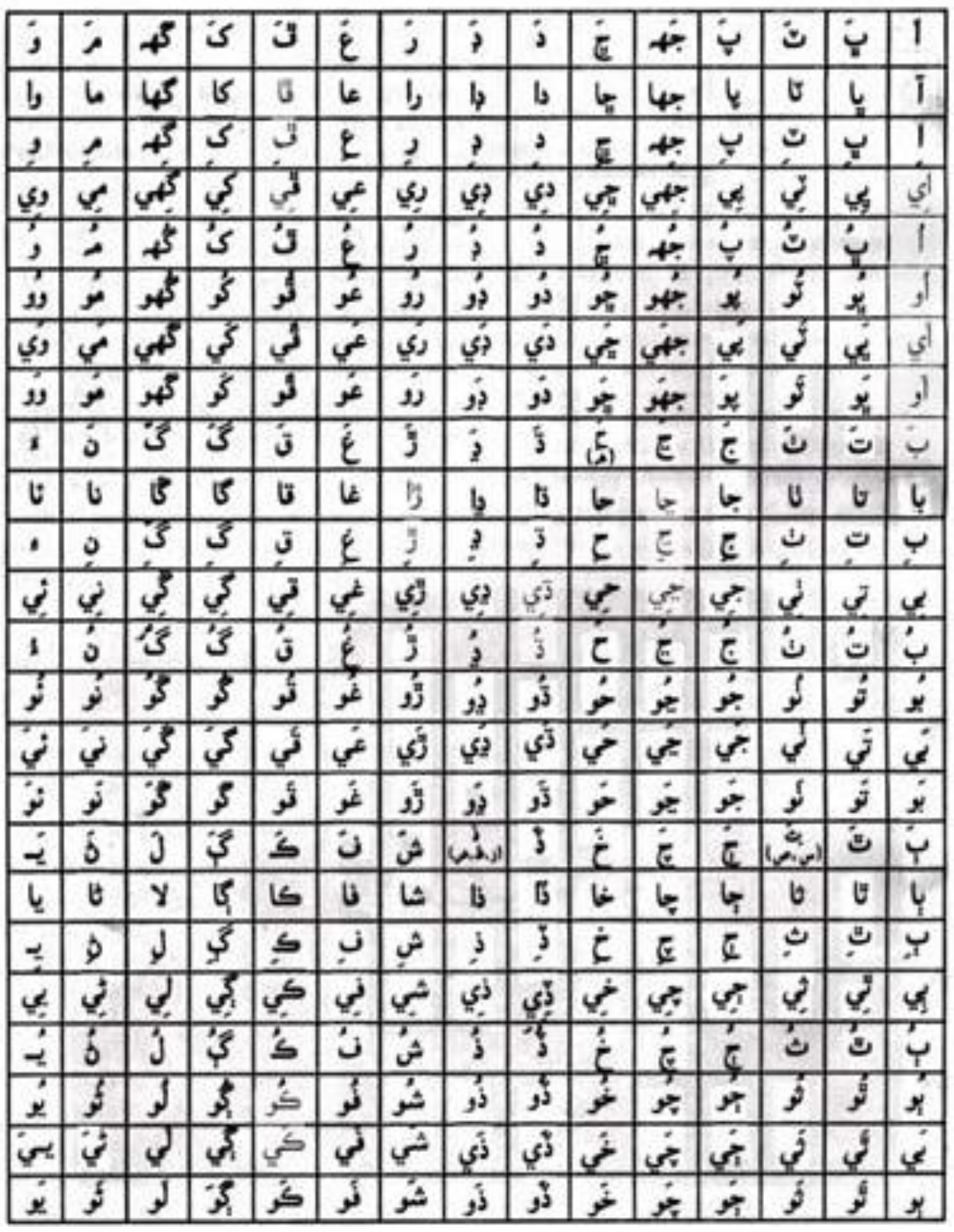

**Fig. 1:** List of 360 Sindhi Phonemes

*Generating the Suggestions List:*

In every spell checker application, the major responsibility is to be able to generate a suitably accurate suggestion list for the misspelled word. Therefore, in order to develop a Sindhi spell checker system, we first need to work out how the suggestion list will be generated. In order to develop a procedure or routine for finding similar words list, following simple architecture has been implemented shown in figure 2 consisting of three main algorithms. According to this architecture the three algorithms that will be used in the system are Levenshtein Edit Distance algorithm (also commonly known as Edit Distance algorithm), second is the Sindhi phonetic based SoundEx algorithm and the last algorithm used is the ShapeEx pattern matching algorithm. The Levenshtein algorithm has not been discussed in this paper as already huge literate is present on the subject and currently is irrelevant to the context of this paper. Both phonetic algorithms have been developed for Sindhi Language and are implemented to generate similar words for any given word. Then the generated list of similar words form each is combined and redundant words are removed. These two algorithms are further discussed in details in the following sections.

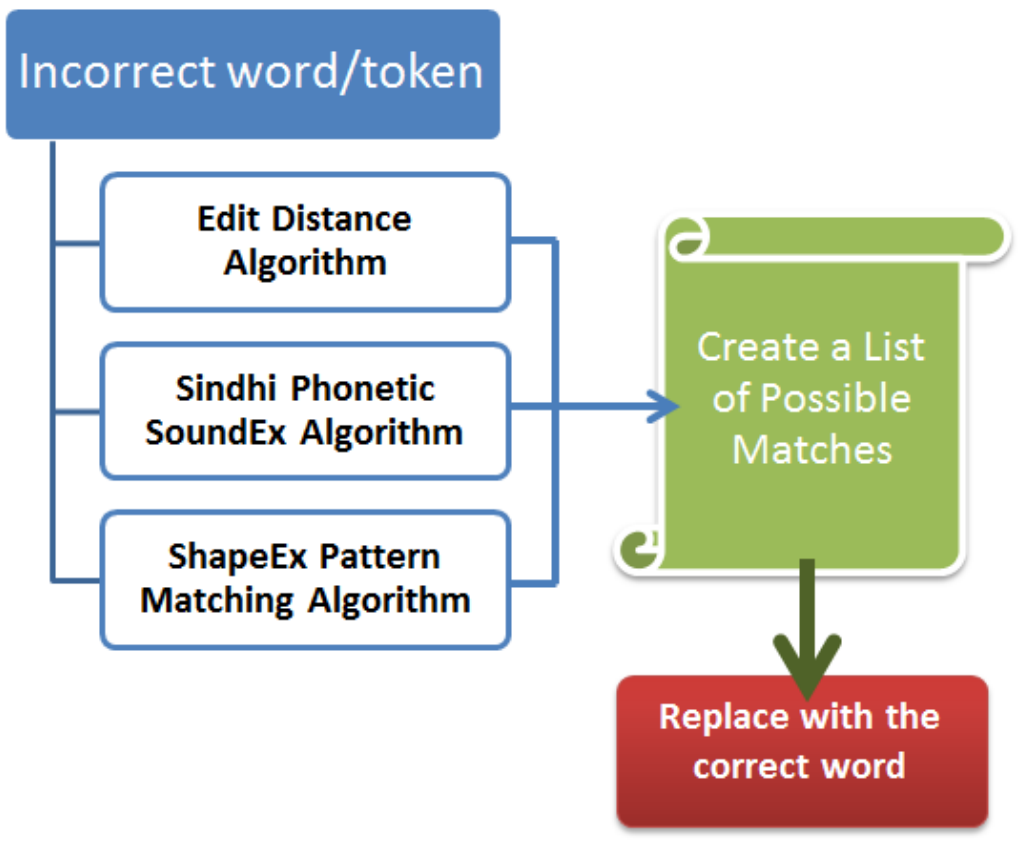

**Fig. 2:** Basic architecture of generating the suggestions

*SoundEx Algorithm:*

The first algorithm used for generating similar words suggestion list is the SoundEx phonetic algorithm, which is used to encode the words and character by their sound or pronunciation. According to this algorithm each characters or letter in a language is assigned a number or code and the characters with similar homophones are all grouped together with same code so that when the comparison is done the words with similar homophones characters, can be matched regardless of minor differences in spelling. For Sindhi language the character haves been grouped according to the similar homophones of each characters as shown in the table 1. All the letters in Sindhi language have been grouped and an alpha-numeric code has been assigned to each group. There also exist few characters in Sindhi language that can't be grouped with any other characters as their pronunciation is completely different from the rest of the characters. There are 22 groups that have been created for Sindhi.

**Table 1:** Sindhi SoundEx based Character groups

| Sound code | what Sound Character Group Similar | | | | | | S.No |
|---|---|---|---|---|---|---|---|
| 0 | | ئ | ي | ء | آ | ا | 1. |
| 1 | | | | | | پ | 2. |
| 2 | | | | ٻ | ب | ڀ | 3. |
| 3 | | | | | ط | ت | 4. |
| 4 | | | | | | ٿ | 5. |
| 5 | | | | | ٺ | ٽ | 6. |
| 6 | | | | ص | س | ث | 7. |
| 7 | | | | چ | ڇ | ج | 8. |
| 8 | | | | | ڃ | ڄ | 9. |
| 9 | | | | ه | ھ | ح | 10. |
| A | | | | | ک | خ | 11. |
| B | | | ڊ | ڏ | ڌ | د | 12. |
| C | | | ظ | ض | ز | ذ | 13. |
| D | | | | ر | ڑ | ڙ | 14. |
| E | | | | | | ش | 15. |
| F | | | | | ڦ | ف | 16. |
| G | | | | | ڪ | ق | 17. |
| H | | | | | غ | گ | 18. |
| I | | | | | | ل | 19. |
| J | | | | | | م | 20. |
| K | | | | | | ن | 21. |
| L | | | | | ڱ | ڳ | 22. |

According to the soundEx algorithm for Sindhi; to find a similarly pronounced word for a given misspelled word, first the soundEx based code is generated for the misspelled word, then that code is compared with the soundEx of each word from the correctly spelled Sindhi words repository. Hence the matching of string is done using the soundEx code not the actual Sindhi text string, thus increasing the efficiency. All the words that have been matched are then put into the similar word list. We used a simple vector container in java to hold this words list.

The comparison according to the Sindhi soundEx algorithm is shown in table 2, where two Sindhi words are shown with similar soundEx code. The two words used here are پاڪصتان{Pakestan: noun} which is an incorrect or misspelled word and the correct spelling of this word is پاڪستان{Pakistan: noun}. Since currently, the only error here is the letter ص{~Swad} in misspelled word, which should be replaced with س{~Sceen}. Both these letter fall into same soundEx group as the pronunciation of both have same phonology and are similar to also in phonetics. As they are in same group both have same numerical code, this allows us to easily find letters that have same code using table 1. But in context of Spell checker we need entire word therefore the entire word is encoded according to the soundEx table grouping. For the word in table 2, it can be seen that both the words generate same soundEx code, hence the generating of similar words become quite easy and efficient.

**Table 2:** SoundEx comparison of two Sindhi words

| Wrong word | SoundEx Code | Correct Word |
|---|---|---|
| پ | 1 | پ |
| ا | 0 | ا |
| ڪ | G | ڪ |
| ص | 6 | س |
| ت | 3 | ت |
| ا | 0 | ا |
| ن | K | ن |
| پاڪصتان | 10G630K | پاڪستان |

Using this soundEx algorithm we were able to generate approximately 43,627 similar words list for Sindhi Language. We used Sindhi Dictionary containing more than 100,000 words, using the technique discussed in [29]. Table 3 shows some of the generated similarly pronounced Sindhi words using phonetic soundEx algorithm. All the similar sounding words generated, have same soundEx code and thus were grouped.

**Table 3:** List of Similar words generated using SoundEx Algorithm

| S.No | Sindhi Similar Words List | SoundEx code |
|---|---|---|
| 1. | ٻپر , ٻپڙ , ٻپڻ , آبر , آبڙ , آبڻ , آپر , آپڻ , ابر , اٻر , اٻڙ , اپر , اپڻ | x02DZ |
| 2. | ٻپا , ٻپا , ٻپي , آبا , آبي , آپا , آپي , ابا , ابي , اپا , اپي , ٻبا , ٻبي | x020Z |
| 3. | وڍر , وڏر , وڌڙ , وڌڻ , وڍڻ , وڏڙ , وڏڻ | xMBDZ |
| 4. | وٽي , ورء , ورئ , ورا , وري , وڙا , وڙي | xMD0Z |
| 5. | رايون , ريئون , ڙائون , ڙايون , رائون | xD00MKZ |
| 6. | سايا , سايع , سايي , سعيا , سعيي , سيئا , سياء , سيائ , سيئي , سييا | x6000Z |
| 7. | جراع , جڙيء , چڙائ , چڙئي , چڙاي , چرائ , چڙيء , چٽيا , جٽئي , چٽائ | x7D00Z |
| 8. | ٻرڙا , ٻرٽي , برڙي , برٽي , پرڙا , پرڙي , پرٽي | x2DD0Z |
| 9. | پينگي , ٻانگي , پانگا , ٻانگا , ٻانگي | x20KH0Z |
| 10. | ڱاٽا , ڱاٽي , ڱيري , ڳارا , ڳاري , ڳاڙا , ڳاڙي , ڳاٽي , ڳيرا , ڳيري , ڳيڙء , ڳيڙا , ڳيٽا , ڱارا , ڱاري | xL0D0Z |

*ShapeEx Algorithm:*

The second algorithm used in the Sindhi spell checking system to generate the similar words suggestion list is the ShapeEx algorithm, which is used to encode the words and character by similarity of the shape and glyph hence also termed as glyph matching algorithm. Similar to the soundEx algorithm, each characters or letter in Sindhi script is assigned a number or code and all the characters with similar shapes or glyphs are all grouped together with same code so that when the comparison is done the words with similar shape characters, can be matched regardless of minor differences in spelling. For Sindhi language the character haves been grouped according to the similarity of the shapes or glyph of each characters as shown in the table 4. All the letters in Sindhi language have been grouped and a simple alpha-numeric code has been assigned to each group. There are 18 groups that have been created for Sindhi language.

**Table 4:** ShapeEx based character groups

| Shape Code | Similar Shaped Character Groups | | | | | | S.No |
|---|---|---|---|---|---|---|---|
| 0 | | إ | أ | ل | ا | آ | |
| 1 | | | ڀ | پ | ٻ | ب | |
| 2 | | ث | ٺ | ٽ | ٿ | ت | |
| 3 | | ڇ | چ | ڃ | ڄ | ج | |
| 4 | | | | | خ | ح | |
| 5 | | | ذ | ڏ | ڌ | د | |
| 6 | | | | | ڊ | ڍ | |
| 7 | | و | م | ز | ڙ | ر | |
| 8 | | | | | ش | س | |
| 9 | | | | | ض | ص | |
| A | | | | | ظ | ط | |
| B | | | | | غ | ع | |
| C | | | | ق | ڦ | ف | |
| D | | | | | | ڪ | |
| E | | | ڱ | ڳ | گ | ک | |
| F | | | | | ڻ | ن | |
| G | | | | | ه | ھ | |
| H | | | | ء | ئ | ي | |

According to the ShapeEx algorithm for Sindhi, to find a similarly shaped word for a given misspelled word, first the shapeEx based code is generated for the misspelled word, then that code is compared with the shapeEx code of each word from the correctly spelled Sindhi words list, and all the words that matched, are then put into the similar word list. The comparison according to the Sindhi shapeEx algorithm is shown in table 4 below where two Sindhi words are shown with similar shapeEx code.

**Table 4:** ShapeEx comparison of two Sindhi words

| **Wrong word** | **ShapeEx Code** | **Correct Word** |
|---|---|---|
| ٻ | 1 | ب |
| ڪ | D | ڪ |
| ر | 7 | ر |
| ي | H | ي |
| بڪري | 1D7H | ٻڪري |

Similar to the soundEx method, SoundEx algorithm yielded approximately 49,071 similar words table for Sindhi Language. Here again we used the same Sindhi Dictionary containing more than 100,000 words in its repository. Table 5 shows some of the generated Sindhi words with similar shape features, using shapeEx algorithm. All the similar shape words generated have same shapeEx code and thus were grouped.

**Table 5:** List of similar words generated using ShapeEx Algorithm

| **S.No** | **Sindhi Similar Words List using** | **ShapeEx Code** |
|---|---|---|
| 1. | اسامي , اسامه , اساري , آسامي , آساري , اشاري , اشاره , لشاري | x0807ZZ |
| 2. | وڱر , مڱر , مگر , زگر , رڳڙ , رڳو , رڳر , ڙڳو , رگو , ڙگڙ , رگو , رگز , رگر , وڳڙ , وڳو , وڳر | x7E7Z |
| 3. | ارهو , ارهر , اڙهو , لوهڙ , لوهو , لوهر , لڙهو , آزهر , آرهڙ , اوهڙ , اوهو , اوهر , ازهر | x07G7Z |
| 4. | زمائي , مزائي , رڙائي , رمائي , روائي , موليٰ , موليه , مرائي , ورائي , وڙائي , ومائي , | x770ZZ |
| 5. | مرون , مرمن , وروڻ , ورون , وڙون , ورمن , ورزن , روڙڻ , رومڻ , رومن , روزن , موڙڻ , موڙن , رمون , رمزن , مومن , ڙمون , مورڻ , مورن , ممون , ووڙڻ , وورڻ , | x777FZ |
| 6. | ومري , رزمي , مروّ , مروه , مرمي , ورڙي , وزمي , ورومئ , ورمي , ورزي , ورري , زومي , زوري , زمره , زرڙي , روڙي , رومي , روزي , روزه , روره , موڙي , رمزي , مومي , موزي , موزه , موري , ووڙي , ومڙي , | x777ZZ |
| 7. | ڳول , ڳوا , ڳرا , ڱڙا , ڱرا , گول , گوا , گزل , گرا , ڳڙا , ڱڙا , | xE70Z |
| 8. | ڳوه , ڳوئ , ڳمي , ڳري , ڱڙي , ڱري , گوي , گوه , گوئ , گزي , گري , گره , ڳڙي , ڳڙه , ڱڙي | xE7ZZ |
| 9. | کرايل , ڪوليا , ڪرايل , ڪرايا , ڪمايل , کوليل , | xD70Z0Z |
| 10. | کنجڙ , کنجو , کنجر , ڪٽچو , ڪنچر , ڪنجر , ڪنجو , ڪنجر , کٽجو | xDF37Z |

*Results:*

The Phonetic algorithms, SoundEx and ShapeEx are most commonly used algorithms in spell checkers for English and other regional languages. The previous work done in this regard for Sindhi language is non-existent. Therefor two tables were initially created defining the grouping of Sindhi letters based on SoundEx and ShapeEx principles. We then developed a test application to examine our results by implementing these two algorithms as shown in figure 3. The application uses an internal Sindhi keyboard which is directly mapped to English keyboard; hence the user is freed from the boredom of installing and setting up Regional languages settings; as discussed by Bhatti et al., [30], to enter the Sindhi text without even installing the additional Sindhi keyboard. As the user enters the text in the application all the incorrect or misspelled words are flagged with red wavy line. The user right clicks the marked word and the system provides the list of possibly correct suggestions for the misspelled word [31].

Initially we test the generation of words list using soundEx algorithm, hence all the words that have been generated using soundEx are appended with "SOX ::". Figure 3 shows the GUI of the system where the image on the left has a huge list of similar words whereas the image on right has only one similar word for the misspelled encircled Sindhi word. This also shows the robustness of the system.

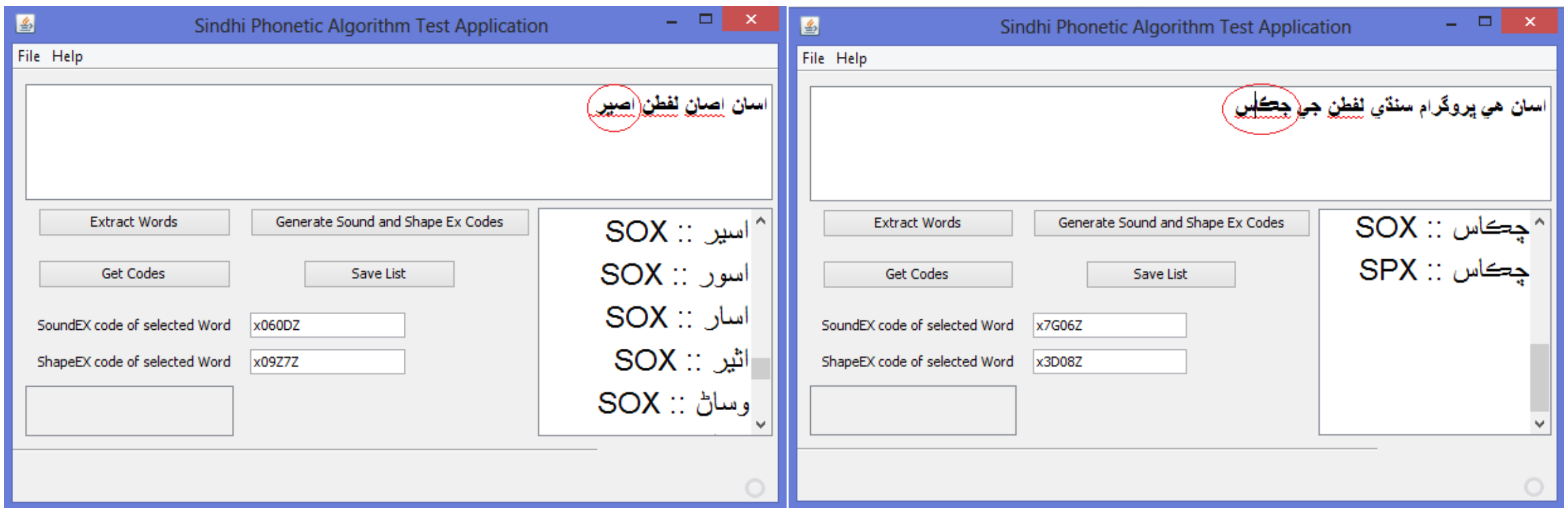

**Fig. 3:** GUI showing the Words list generated using the SoundEx algorithm marked with SOX::{Sindhi Word}

For the image on the left in figure 3, the soundEx code is "x060DZ". Here the character 'x' is appended to identify that it's and encoded string. The full list of words generated for the misspelled word using soundEx algorithm is given below:

SOX :: اسوڙ
SOX :: اسير
SOX :: اسور
SOX :: اسار
SOX :: اثير
SOX :: وساڻ
SOX :: وسار
SOX :: آسير
SOX :: آساڙ
SOX :: آسار
SOX :: يسوڻ
SOX :: يسير
SOX :: آثار
SOX :: يساڻ
SOX::يسار

The ShapeEx algorithm was then used to generate the list and all the words generated using this algorithm were appended with "SPX ::" for identification purpose. Figure 4 shows the GUI of the system where the misspelled words is marked with red line, and upon user interaction a suggestion list is displayed below using the shapeEx algorithm. Again it was noted that the number of suggestions generated, varied according to the misspelled word from one correct word to multiple.

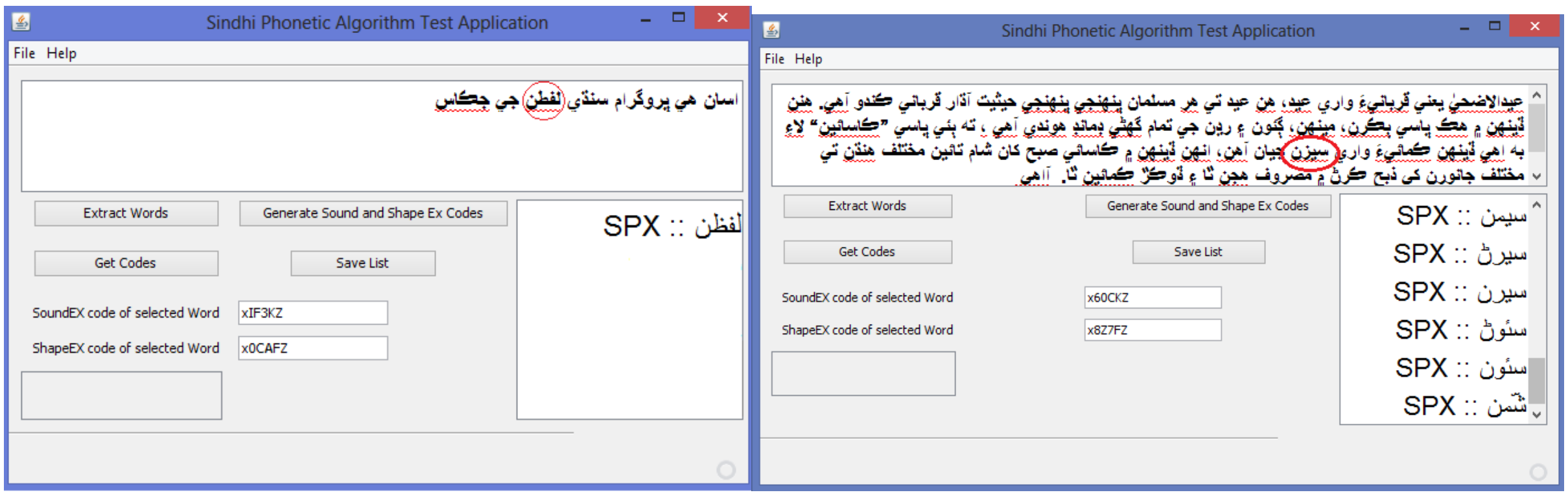

**Fig. 4:** GUI showing the Words list generated using the ShapeEx algorithm marked with SPX::{Sindhi Word}

For the word with shapeEx code of "x0ZZZ" following list of correctly spelled words is generated.
SPX word is::لٰه
SPX word is::لّي
SPX word is::لّه
SPX word is::لّة
SPX word is::لٰي
SPX word is::ليّ
SPX word is::ليه
SPX word is::لية
SPX word is::ليئ
SPX word is::لئي
SPX word is::ايّ
SPX word is::ائي
SPX word is::اً;
SPX word is::آيه
SPX word is::آيئ

By combining both these algorithms, richly populated list of correctly spelled similar Sindhi words are obtained that can be easily used to replace the misspelled word in any Sindhi Text application as shown in figure 5. Obviously some of the suggested word in the list don't exactly match with the misspelled word and need refinement but this problem is left open for further research and actual implementation of Spell checker.

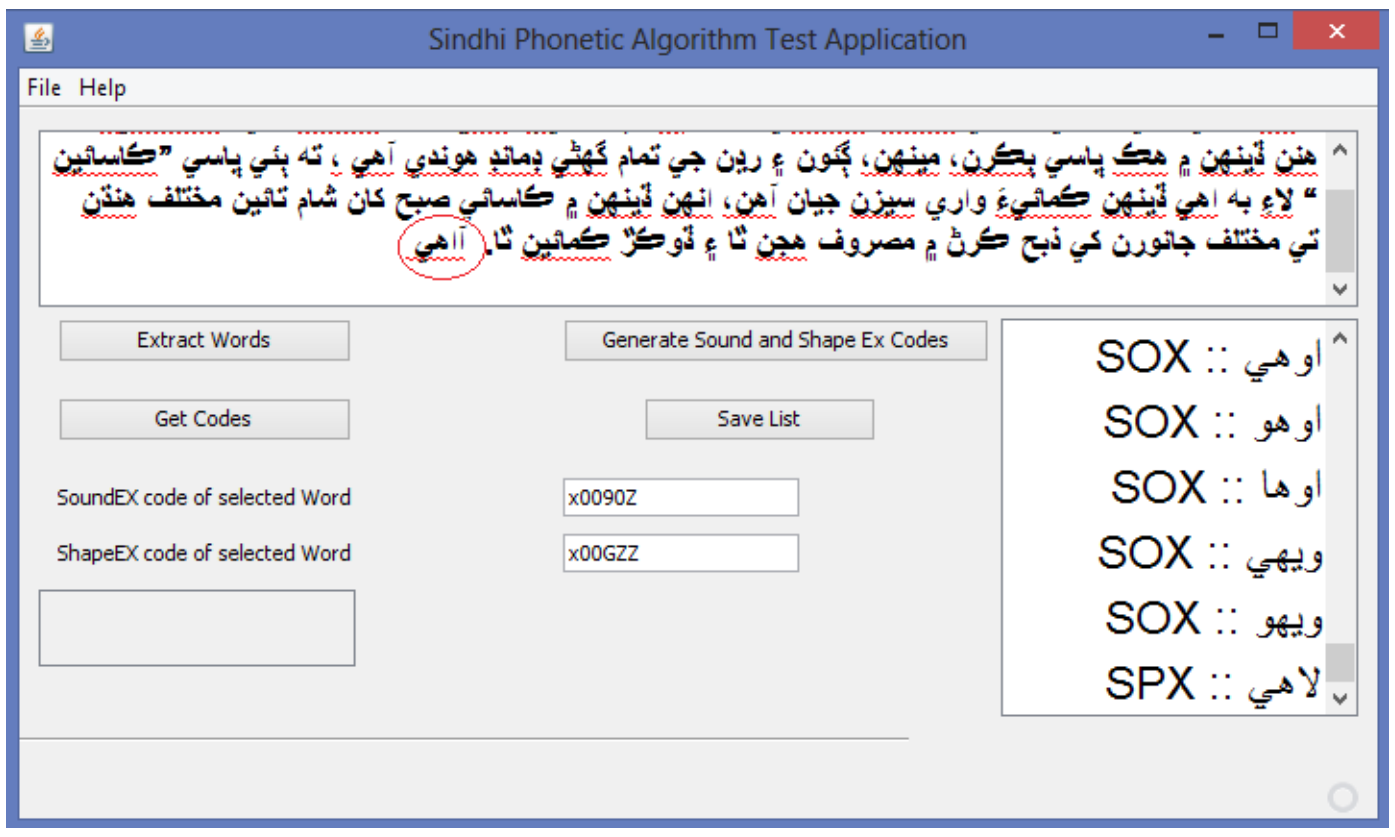

**Fig. 5:** GUI showing the Words list generated using both algorithms

By further refining the two algorithms along with Edit Distance algorithm, the suggestion list of words can be compressed to be more accurate and concise for the final Sindhi Spell checker product as shown in figure 6. This Sindhi spell checker system then can be further utilized and integrated in other online Sindhi Systems under development like [32].

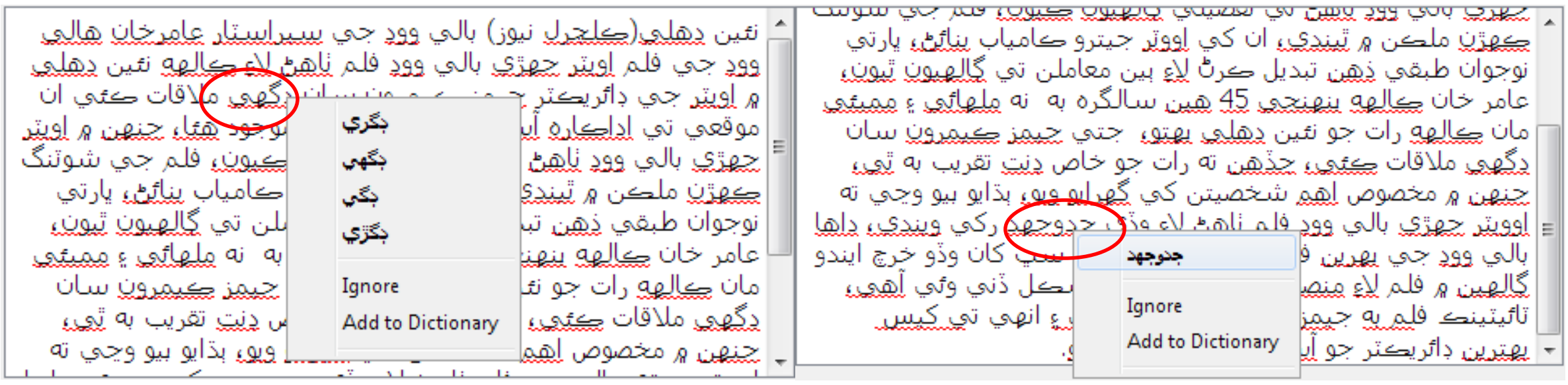

**Fig. 6:** Sindhi Text with consolidated suggestions list for a misspelled word

### *Conclusion and Future Work:*

In this paper we discussed the first ever attempt made to solve the problem of developing a Spell checker for Sindhi language, using phonetic based Sindhi SoundEx and ShapeEx algorithms. Sindhi being a very rich and old language has vast plethora of written text but unfortunately no previous attempt had been made to create such a system; as per best of our knowledge. We have used a combinational approach to generate a similar words list for the misspelled Sindhi string. A phonetic based Sindhi SoundEx algorithm was developed along with a pattern or glyph matching algorithm called ShapeEx. For these two algorithms, first time ever the Sindhi Phonetic based grouping was done for Sindhi characters based on their pronunciation and secondly based on their shapes or glyph representations. The system is able to identify the misspelled words correctly and then provides the suggestion list very accurately when desired by the user for the correction of the word. In future we would like to extend the SoundEx algorithm to Metaphone and Double-Metaphone based algorithm for higher degree of accuracy and efficiency.

## REFERENCES

[1]. Kukich, K., 1992. Techniques for automatically correcting words in text, in .ACM Computing Surveys., 24 (4): 377-439

[2]. Zobel, J. and P. Dart, 1995. “Finding Approximate Matches in Large Lexicons”, Software- Practice and Experience, 25(3): 331-345.

[3]. Hull, D.A., 1996. Stemming algorithms: a case study for detailed evaluation. Journal of the American Society for Information Science, 47(1): 70-84.

[4]. Hodge, V.J., J. Austin & Y. Dd, 2001. An evaluation of phonetic spell checkers. In Mechanisms of Radiation Eflects in Electronic Materials.

[5]. Segalovich, Ilya., 2003. "A fast morphological algorithm with unknown word guessing induced by a dictionary for a web search engine." Proceedings of MLMTA.

[6]. Bhatti, Z., I.A. Ismaili, A.A. Shaikh, And W. Javaid, 2012. Spelling Error Trends and Patterns in Sindhi, Journal of Emerging Trends in Computing and Information Sciences, 3(10): 2079-8407.

[7]. NASEEM, T., S. HUSSAIN, 2007. “Spelling Error Trends in Urdu”. In the Proceedings of Conference on Language Technology (CLT07), University of Peshawar, Pakistan.

[8]. Raghavan, H., & J. Allan, 2004. Using soundex codes for indexing names in ASR documents. In Proceedings of the Workshop on Interdisciplinary Approaches to Speech Indexing and Retrieval at HLT -NAACL 2004 (pp. 22-27). Association for Computational Linguistics.

[9]. Jacobs, J.R., 1982. Finding words that sound alike. The Soundex algorithm. Byte, 7(3): 473-474.

[10]. Phillips, L., 1990. “Hanging on the Metaphone”, Computer Language, 7(12): 1990.

[11]. Phillips, L., 2000. “The Double Metaphone Search Algorithm”, C/C++ Users Journal, 18(6), June, 2000. Also available online at http://www.cuj.com/documents/s=8038/cuj0006philips/

[12]. Gadd, T.N., 1990. PHONIX: The algorithm. Program: electronic library and information systems, 24(4): 363-366.

[13]. Jatoi, A.N., 1968. "IlmLisanAin Sindhi Zaban", Institute of Sindhalogy, Hyderabad

[14]. Bugio, M.Q., 2001. "Sociolinguistics of Sindh", LINCOM EUROPA? PhD Thesis.

[15]. Cole, Jennifer, 2006. "The Sindhi Language", In K.Brow(ed) Encyclopedia of Language and Linguistics, 2nd Edition, 11: 384-386. Oxford: Elsevier.

[16]. Cole, Jennifer., 2005. "Sindhi", In Strazny, Philipp(ed) Encyclopedia of Linguistic. New Yark: Routledge.

[17]. Bird, S., 1991, "A Logical Approach To Arabic Phonology", Proceedings of the 5th Conference on European Chapter of the Association for Computational Linguistics, pp: 89-94.

[18]. Zemirli, Z., S. Khabet, M. Mosteghanem, 2007. "An effective model of streesing in an Arabic Text to Speech System", IEEE AICCSA, pp: 700-707.

[19]. Shaalan, K., A. Allam, & A. Gomah, 2003. Towards automatic spell checking for Arabic. In Proceedings of the 4th Conference on Language Engineering, Egyptian Society of Language Engineering (ELSE) pp: 240-247.

[20]. Source: http://company.yandex.ru/articles/iseg-las-vegas.xml

[21]. Ousidhoum, N.D., & N. Bensaou, 2013. Towards the Refinement of the Arabic Soundex. In Natural Language Processing and Information Systems, pp: 309-314). Springer Berlin Heidelberg

[22]. UzZaman, N., & M. Khan, 2005. A double metaphone encoding for Bangla and its application in spelling checker. In Natural Language Processing and Knowledge Engineering, 2005. IEEE NLP-KE'05. Proceedings of 2005 IEEE International Conference on pp: 705-710). IEEE.
[23]. Krishna, B., P. Shrestha, 2007. "Nepali Spellchecker", PAN Localization Working Papers 2007, Centre for Research in Urdu Language Processing, National University of compute and Emerging Sciences, Lahore, Pakistan, pp: 316-318.
[24]. Dhanabalan, T., R. Parthasarathi, & T.V. Geetha, 2003. Tamil Spell Checker. In Sixth Tamil Internet 2003 Conference, Chennai, Tamilnadu, India.
[25]. Naseem, T. And S. Hussain, 2007a. "A Novel Approach for Ranking Spelling error corrections in Urdu", Language Resources and Evaluation. 41.2, p.p:117–128, DOI 10.1007/s10579-007-9028-6, © Springer Science+Business Media B.V.
[26]. Naseem, T., & S. Hussain, 2007b. Spelling Error Corrections for Urdu . Published online: 26 September 2007 © Springer Science+Business Media B.V. 2007. PAN Localization Working Papers 2007, Centre for Research in Urdu Language Processing, National University of compute and Emerging Sciences, Lahore, Pakistan, pp: 117-128.
[27]. Kaur, R., P. Bhatia, 2010. Design and Implementation of SUDHAAR-Punjabi Spell Checker. International Journal of Information and Telecommunication Technology (IJITT), 1(1) (ISSN: 0976-5972), North America.
[28]. Mahar, J.A., & G.Q. Memon, 2009. Phonology for Sindhi Letter to Sound Conversion. Journal of Information and Communication Technology, 3(1): 11-20.
[29]. Shah, A.A., A.W. Ansari, & L. Das, 2004. Bi-Lingual Text to Speech Synthesis System for Urdu and Sindhi. In National Conf. on Emerging Technologies, pp: 20126-130.
[30]. Ismaili, I.A., Z. Bhatti, A.A. Shah, 2012. Development of Unicode based bilingual Sindhi-English Dictionary. Mehran University Research Journal of Engineering & Technology, 31(1) [ISSN 0254-7821].
[31]. Bhatti, Z., I.A. Ismaili, W.I. Khan, A.S. Nizamani, 2013a. Development of Unicode based Sindhi Typing System, Journal of Emerging Trends in Computing and Information Sciences, 4(3): 309-314, ISSN 2079-8407
[32]. Bhatti, Z., I.A. Ismaili, W.J. Soomro, & D.N. Hakro, 2014. Word Segmentation Model for Sindhi Text. American Journal of Computing Research Repository, 2(1): 1-7.DOI: 10.12691/ajcrr-2-1-1
[33]. Bhatti, Z., D.N. Hakro, & A.A. Jarwar, 2013b. Sindhi Academic Informatic Portal. American Journal of Information Systems, 1(1): 21-25. DOI: 10.12691/ajis-1-1-3.